%% file: main.tex
\documentclass[acmtog]{acmart}

\AtBeginDocument{%
  }

\setcopyright{none}

\input{preamble}

\begin{document}

\title{GeoQuery: Geometry-Query Diffusion for Sparse-View Reconstruction}

\author{Xiao Cao}
\authornote{This work was done during Xiao Cao's internship at Rawmantic AI.}
\affiliation{%
  \institution{University of Electronic Science and Technology of China \& Rawmantic AI}
  \city{Chengdu}
  \country{China}
}
\email{xiaocao@std.uestc.edu.cn}

\author{Yuze Li}
\affiliation{%
  \institution{Tianjin University}
  \city{Tianjin}
  \country{China}
}

\author{Youmin Zhang}
\affiliation{%
  \institution{Rawmantic AI}
  \city{Chengdu}
  \country{China}
}

\author{Jiayu Song}
\affiliation{%
  \institution{Rawmantic AI}
  \city{Chengdu}
  \country{China}
}

\author{Cheng Yan}
\affiliation{%
  \institution{Tianjin University}
  \city{Tianjin}
  \country{China}
}

\author{Wen Li}
\authornote{Corresponding author.}
\affiliation{%
  \institution{University of Electronic Science and Technology of China}
  \city{Chengdu}
  \country{China}
}
\email{liwenbnu@gmail.com}

\author{Lixin Duan}
\affiliation{%
  \institution{University of Electronic Science and Technology of China}
  \city{Chengdu}
  \country{China}
}

\input{main/00_abstract}


\begin{teaserfigure}
  \includegraphics[width=\textwidth]{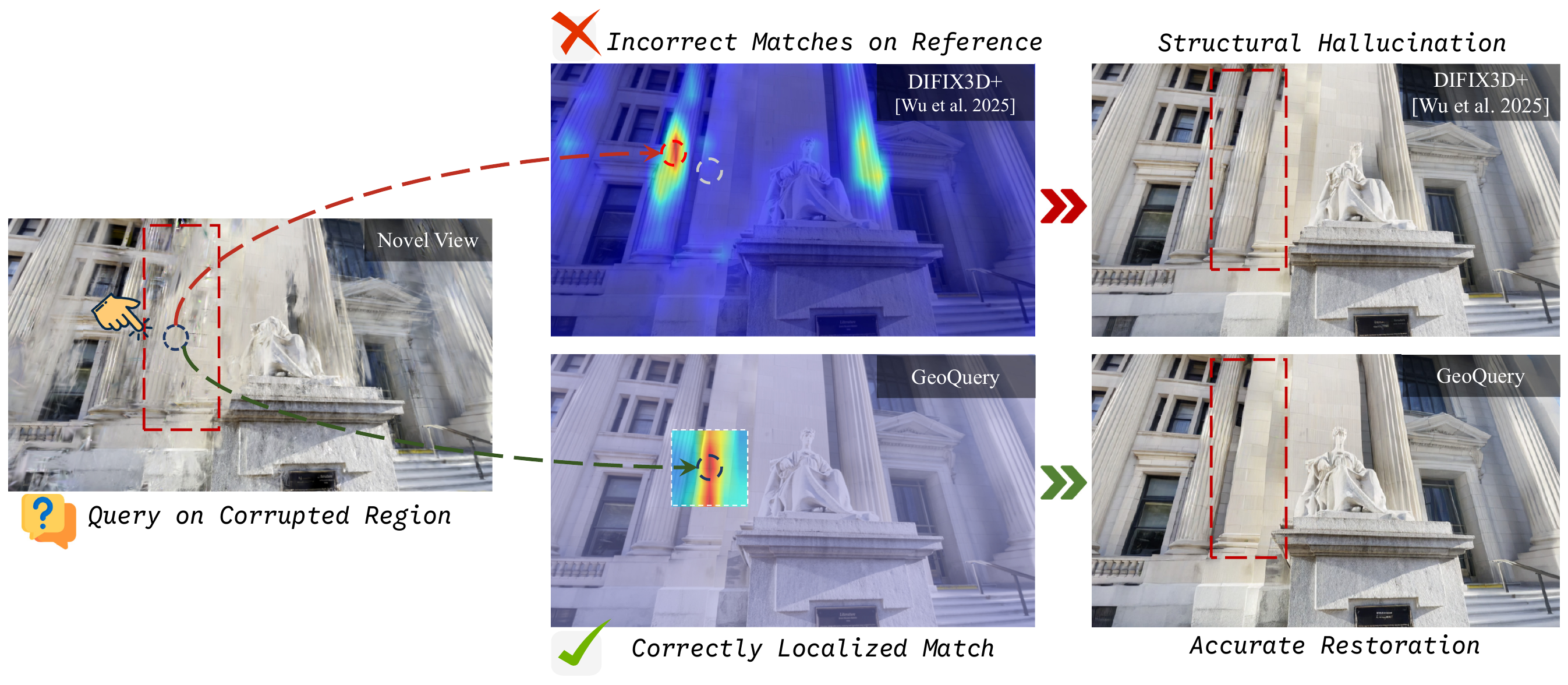}
  \caption{Our method, \textbf{GeoQuery}, enables high-fidelity restoration of corrupted 3DGS renderings. \textbf{Left:} When a query originates from a corrupted region (highlighted by the dashed box), standard multi-view attention (Top, DIFIX3D+) suffers from query contamination, retrieving incorrect matches scattered across the reference view. This semantic misalignment leads to severe structural hallucinations in the final output. \textbf{Bottom:} In contrast, GeoQuery leverages geometry-induced correspondences to enforce a correctly localized match (cyan box), strictly anchoring feature retrieval to the physical geometry. This enables the precise transfer of clean details from the reference, resulting in an accurate restoration that preserves structural integrity.}
  \label{fig:teaser}
\end{teaserfigure}

\maketitle
\input{main/01_intro}

\input{main/02_related_works}

\input{main/03_preliminary}

\input{main/04_methods}
\input{main/05_experiment}

\input{main/06_conclusion}

\begin{acks}
This work is supported by the New Generation Artificial Intelligence-National Science and Technology Major Project (No. 2025ZD0123002).
\end{acks}

\bibliographystyle{ACM-Reference-Format}
\bibliography{sample-base}

\end{document}

%% file: preamble.tex
\usepackage{colortbl}
\usepackage{graphicx}
\usepackage{subcaption}
\usepackage{xspace}
\usepackage{multirow}
\usepackage{makecell}
\usepackage{algorithm}
\usepackage{booktabs}
\usepackage{pifont}
\usepackage{stfloats}

\newcommand{\best}{\cellcolor{tablered}}
\newcommand{\sbest}{\cellcolor{orange}}

\makeatletter
\DeclareRobustCommand\onedot{\futurelet\@let@token\@onedot}
\def\@onedot{\ifx\@let@token.\else.\null\fi\xspace}

\makeatother

\definecolor{yellow}{rgb}{1, 1, 0.7}
\definecolor{orange}{rgb}{1, 0.85, 0.7}
\definecolor{tablered}{rgb}{1, 0.7, 0.7}
\definecolor{red}{rgb}{1, 0, 0}

\definecolor{wincolor}{rgb}{0.85, 0.0, 0.0}

\definecolor{darkyellow}{rgb}{0.8, 0.8, 0.5}
\definecolor{darkred}{rgb}{0.7, 0.3, 0.3}
\definecolor{darkgreen}{rgb}{0.3, 0.7, 0.3}
\definecolor{green}{rgb}{0, 1.0, 0}
\definecolor{pink}{rgb}{1, 0.4, 0.7}



%% file: main/00_abstract.tex
\begin{abstract}
3D Gaussian Splatting (3DGS) has emerged as a prominent paradigm for 3D reconstruction and novel view synthesis. However, it remains vulnerable to severe artifacts when trained under sparse-view constraints. While recent methods attempt to rectify artifacts in rendered views using image diffusion models, they typically rely on multi-view self-attention to retrieve information from reference images. We observe that this mechanism often fails when the rendered novel views output by 3DGS are heavily corrupted: damaged query features lead to erroneous cross-view retrieval, resulting in inconsistent rendering refinement. To address this, we propose \emph{\textbf{GeoQuery}}, a geometry-guided diffusion framework that integrates generative priors with explicit geometric cues via a novel Geometry-guided Cross-view Attention (GCA) mechanism. First, by leveraging predicted depth maps and camera poses, we construct a geometry-induced correspondence field to sample reference features, forming a geometry-aligned proxy query that replaces the corrupted rendering features. Furthermore, we design a new cross-view feature aggregation pipeline, in which we restrict the cross-view attention to a local window around each proxy query to effectively retrieve useful features while suppressing spurious matches. GeoQuery can be seamlessly integrated into existing diffusion-based pipelines, enabling robust reconstruction even under extreme view sparsity. Extensive experiments on sparse-view novel view synthesis and rendering artifact removal demonstrate the effectiveness of our approach. Code is available at \href{https://xiaoc7.github.io/GeoQuery/}{\textcolor{pink}{Project Page}}.

\end{abstract}

%% file: main/01_intro.tex
\section{Introduction}

Sparse-view 3D reconstruction and novel view synthesis (NVS) remain longstanding challenges in computer vision and graphics.
While 3D Gaussian Splatting (3DGS)~\cite{kerbl20233d} has enabled real-time high-fidelity rendering, its explicit representation is prone to overfitting under sparse observations, resulting in geometric collapse and floating artifacts in novel views.
To mitigate these issues, recent \emph{render and refine} pipelines ~\cite{hirschorn2025splatent, wu2025difix3d+, wu2025genfusion, yin2025gsfixer, liu20243dgs, wu2024reconfusion} integrate Diffusion Models \cite{ho2020ddpm,song2020sde,rombach2022ldm} to hallucinate missing details and repair artifacts in 3DGS renderings, using the refined images as pseudo observations.
However,  a critical dilemma arises in how these diffusion models utilize reference views. Recent generative refinement approaches \cite{wu2025difix3d+,hirschorn2025splatent} typically facilitate cross-view information exchange via multi-view self-attention \cite{shi2023mvdream,liu2024one2345,liu2023zero123,shi2023zero123++}. In this paradigm, features from the target and reference views are concatenated, allowing the noisy target tokens to attend to reference tokens for context retrieval. While this mechanism captures long-range semantic context effectively, it is inherently unstable when the query derived from the rendering is corrupted by severe artifacts. We term this phenomenon \emph{\textbf{Query Contamination}}. When the rendering contains severe floaters or distortions, the contaminated query retrieves semantically similar but geometrically irrelevant texture from the reference view (e.g., the first row in Fig.~\ref{fig:teaser}). This erroneous retrieval reinforces the artifacts instead of removing them, creating a vicious cycle.

To resolve this problem, we propose \emph{\textbf{GeoQuery}}, a geometry-guided diffusion framework that parallels the global semantic search with a geometry-induced retrieval branch. 
Specifically, in parallel to the standard UNet features, we introduce a Geometry-Guided Cross-View Attention (GCA) module that implements a geometry-indexed query substitution mechanism.
Recognizing that features extracted from artifact-corrupted renderings are unreliable, we bypass them to prevent query contamination.
Instead, we leverage the geometric correspondence, derived from estimated depth maps and camera poses, to identify specific tokens in the reference view that are spatially aligned with the target positions. These homologous reference tokens are adopted as proxy queries to initiate attention within their local neighborhoods in the reference feature map.
To enforce structural consistency, GCA restricts retrieval to a correspondence-centric local window, effectively filtering out spurious long-range matches. Finally, a learnable gating mechanism adaptively fuses this geometry-induced evidence into the diffusion backbone. This allows the model to dynamically balance the two streams: relying on global context for general texture synthesis while leveraging GeoQuery to rectify structural errors and suppress hallucinations in artifact-prone regions.

We integrate GeoQuery into a progressive 3DGS refinement pipeline \cite{wu2025difix3d+}. Evaluations on the DL3DV-Benchmark \cite{ling2024dl3dv} and Mip-NeRF360 dataset \cite{barron2022mip} demonstrate that GeoQuery establishes a new state-of-the-art for sparse-view reconstruction and rendering artifact removal. While maintaining competitive perceptual metrics against leading diffusion-based priors (e.g., DIFIX3D+), our method achieves superior PSNR and more consistent refinement by effectively mitigating mismatched cross-view retrieval. Crucially, in the highly ill-posed 3-view regime, GeoQuery prevents the geometric collapse observed in baselines, yielding robust novel view synthesis where prior approaches fail. For rendering artifact removal, GeoQuery outperforms baselines across all quantitative metrics.

In summary, we make the following contributions:
\begin{itemize}
    \item Analysis of Query Contamination: We identify a critical feedback loop in existing solvers where contaminated queries from corrupted renderings mislead the attention mechanism, causing artifact propagation.
    \item The GeoQuery Framework: We propose a geometry-guided diffusion framework featuring Geometry-Guided Cross-View Attention (GCA). By employing a geometry-indexed query substitution strategy, GCA constructs reliable \emph{Proxy Queries} to strictly anchor the generative process to physical correspondences.
    \item Superior Performance: GeoQuery establishes a new state-of-the-art in artifact removal and effectively prevents geometric collapse in challenging sparse-view scenarios.
\end{itemize}

%% file: main/02_related_works.tex
\section{Related Works}
Neural Radiance Fields (NeRF)~\cite{mildenhall2021nerf} and 3D Gaussian Splatting\cite{kerbl20233d} have substantially advanced scene reconstruction and novel-view synthesis (NVS) by enabling photorealistic rendering from posed multi-view observations. 
Despite strong performance under dense view coverage, these methods become under-constrained in sparse-view training or extreme extrapolation. Missing observations and imperfect geometry often lead to floaters, structural blur, and inconsistent textures. Existing works on addressing these issues for sparse novel view synthesis mainly follow two directions, regularization-based and generative-prior-based, which are summarized as follows.

\paragraph{Regularization-based Novel View Synthesis.}
Previous work often augments 3D optimization with constraints beyond purely photometric objectives\cite{zhu2024fsgs,li2024dngaussian,xu2025dropoutgs, park2025dropgaussian, zhang2024corgs, zhao2025selfenblegs, wang2023sparsenerf, somraj2023simplenerf}. 
A representative line of methods introduces external priors to regularize sparse-view optimization, including geometric cues~\cite{zhu2024fsgs, li2024dngaussian, wang2023sparsenerf, deng2022depthnerf, roessle2022densedepthnerf, niemeyer2022regnerf, charatan2024pixelsplat,chen2024mvsplat,zheng2025nexusgs}, semantics predicted by pretrained models~\cite{jain2021dietnerf, xu2025depthsplat}, and frequency-based regularization~\cite{yang2023freenerf, zhang2024fregs}. Overall, they provide complementary priors that improve reconstruction quality and consistency under limited observations. 
Another line of approaches~\cite{somraj2023simplenerf, zhang2024corgs, zhao2025selfenblegs, xu2025dropoutgs, park2025dropgaussian, patle2025ad} does not rely on external geometric priors. Instead, they introduce heuristic rules or explicit regularization to stabilize training and reduce overfitting under sparse views, which improves generalization to novel viewpoints.
While these strategies reduce ambiguity in under-observed regions, their effectiveness hinges on balancing added constraints with photometric supervision. Overly strong constraints over-smooth details, while weak constraints leave residual artifacts.

\paragraph{{Generative-Prior-based Novel View Synthesis}}
Recent progress in generative modeling~\cite{rombach2022ldm,ho2020ddpm,song2020sde,liu2023zero123,shi2023zero123++,long2024wonder3d,sargent2024zeronvs} has enabled strong priors for repairing degraded observations and synthesizing plausible content in unobserved regions, which has been increasingly adopted to improve sparse-view novel view synthesis.
Building on this capability, recent reconstruction pipelines integrate diffusion models to refine rendered pseudo-observations and use the restored views as stronger supervision for updating the underlying 3D representation under limited observations.
ReconFusion~\cite{wu2024reconfusion} represents an early effort in this direction by coupling image diffusion with posed multi-view conditioning to regularize sparse-view reconstruction.
To encourage cross-view coherence, several follow-up works resort to video diffusion models that operate on trajectories or view sequences rendered from the current reconstruction.
3DGS-Enhancer~\cite{liu20243dgs} trains a video diffusion model on large-scale data to repair additional rendered views and distills the restored results back into a low-quality 3DGS representation.
GenFusion~\cite{wu2025genfusion} further constructs an artifact-prone RGB-D video dataset via masking and fine-tunes a video diffusion model for restoration under structured degradations, facilitating subsequent reconstruction refinement.
GSFixer~\cite{yin2025gsfixer} also follows the video-diffusion paradigm for 3DGS and conditions the restoration process on both 2D semantic cues and 3D geometric features to better handle corrupted novel views.
In contrast, DIFIX3D+~\cite{wu2025difix3d+} builds upon an image diffusion model trained for artifact removal and injects the learned prior into reconstruction via periodic distillation and optional inference-time refinement.
Our work is most closely related to DIFIX3D+~\cite{wu2025difix3d+}, and further investigates how to incorporate explicit geometric cues into an image-diffusion formulation for novel-view restoration.

%% file: main/03_preliminary.tex
\section{Preliminary}

\begin{figure*}[t]
  \centering
  \includegraphics[width=\textwidth]{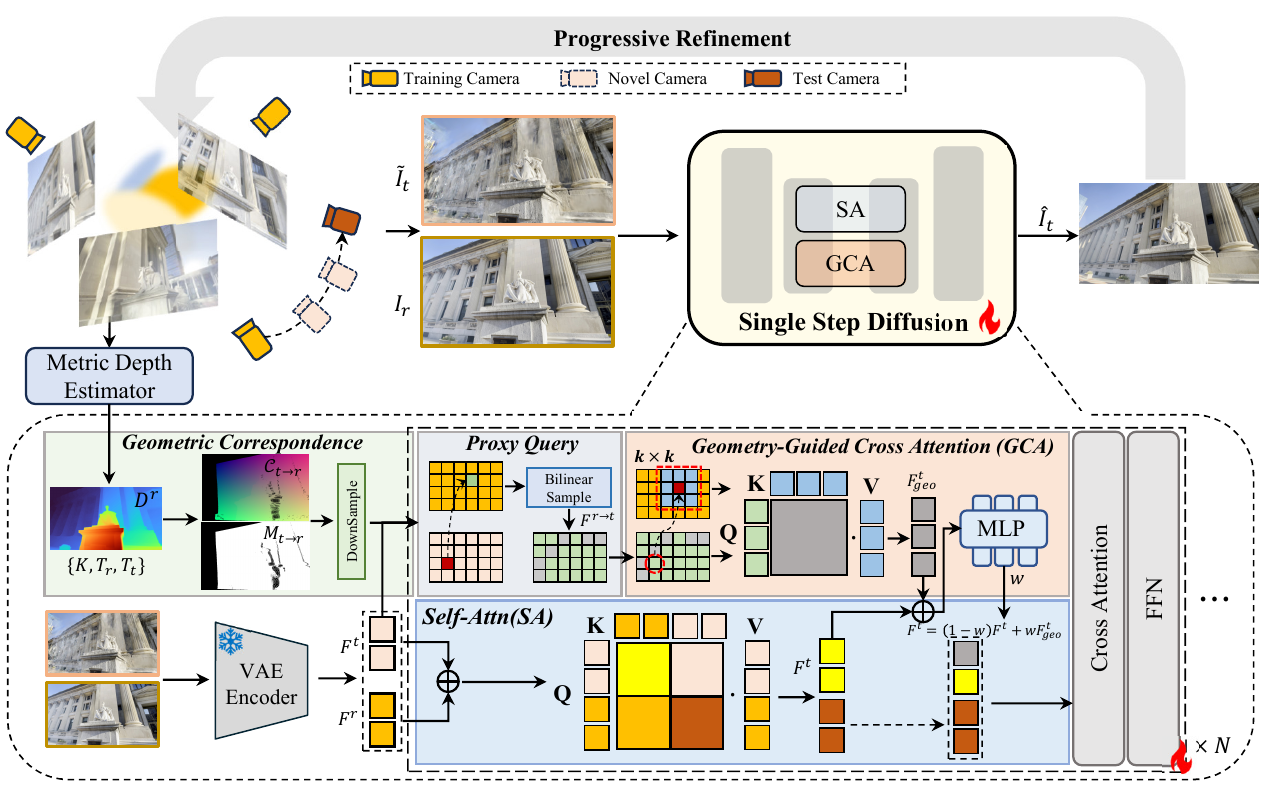}
  \caption{Overview of GeoQuery. Starting from a sparse training set, we optimize a 3D Gaussian Splatting (3DGS) representation and progressively refine it through iterative rendering and supervision updates. At each step, 3DGS produces an artifact-prone rendering $\tilde{I}_t$. We estimate metric depth to construct a geometric correspondence field, which is used to retrieve proxy features from the reference view. The proposed Geometry-Guided Cross-View Attention (GCA) restricts retrieval to a local $k \times k$ neighborhood around the indexed correspondence, and an adaptive fusion module integrates the geometry-guided features into the diffusion backbone. The restored output $\hat{I}_t$ is then used as a pseudo-observation for subsequent 3DGS refinement.}
  \label{fig:GeoQuery-pipeline}
\end{figure*}

\subsection{3D Gaussian Splatting}
3D Gaussian Splatting (3DGS)~\cite{kerbl20233d} represents 3D scenes using a collection of anisotropic Gaussians. Each Gaussian is defined by a center $\mathbf{\mu} \in \mathbb{R}^3$ and a 3D covariance matrix $\Sigma$, with its spatial influence formulated as:
\begin{equation}
G(\mathbf{x}) = \exp\left(-\frac{1}{2}(\mathbf{x}-\mathbf{\mu})^\top \Sigma^{-1}(\mathbf{x}-\mathbf{\mu})\right).
\end{equation}
The covariance is optimized via a factored representation $\Sigma = \mathbf{R}\mathbf{S}\mathbf{S}^\top\mathbf{R}^\top$ to ensure semi-definiteness.

To render the scene, 3D Gaussians are projected into 2D image space. The final color $C$ of a pixel is computed by alpha-blending $N$ ordered Gaussians overlapping the pixel:
\begin{equation}
C = \sum_{i=1}^N c_i \alpha_i G^{2D}_i(\mathbf{x}) \prod_{j=1}^{i-1} (1 - \alpha_j G^{2D}_j(\mathbf{x})),
\end{equation}
where $c_i$, $\alpha_i$, and $G^{2D}_i(\mathbf{x})$ denote the color, opacity, and the evaluation of the $i$-th projected 2D Gaussian at pixel position $\mathbf{x}$, respectively.

\subsection{Diffusion Models}

Diffusion models~\cite{rombach2022ldm,ho2020ddpm,song2020sde} learn a data distribution $p_{\text{data}}(\mathbf{x})$ via iterative denoising.
A forward process progressively perturbs data by adding Gaussian noise:
\begin{equation}
\mathbf{x}_\tau = \alpha_\tau \mathbf{x} + \sigma_\tau \boldsymbol{\epsilon}, 
\quad \boldsymbol{\epsilon} \sim \mathcal{N}(\mathbf{0}, \mathbf{I}),
\end{equation}
where $\tau$ denotes the diffusion timestep.
A neural denoiser $F_\theta$ is trained to predict the target $\mathbf{y}$ (typically the noise $\boldsymbol{\epsilon}$) by minimizing:
\begin{equation}
\mathbb{E}_{\mathbf{x},\tau,\boldsymbol{\epsilon}}
\left[\left\| \mathbf{y} - F_\theta(\mathbf{x}_\tau; \mathbf{c}, \tau) \right\|_2^2 \right],
\end{equation}
where $\mathbf{c}$ represents optional conditioning information.
At inference time, the generative process reverses the noising procedure to recover a clean sample.

%% file: main/04_methods.tex
\section{Method}
\label{sec:method}

Recent sparse-view 3DGS methods improve reconstruction quality by injecting diffusion priors into iterative refinement pipelines \cite{hirschorn2025splatent, wu2025difix3d+, wu2025genfusion, yin2025gsfixer, liu20243dgs}. However, we identify a critical bottleneck in existing multi-view self-attention mechanisms, which we term \textbf{Query Contamination}. When the target rendering $\tilde{I}^t$ contains severe artifacts, the resulting query features become unreliable and often retrieve geometrically inconsistent content from reference views.

To address this issue, we propose \textbf{GeoQuery}, which replaces unreliable target queries with \textit{Geometry-Indexed Proxy Features} sampled from the reference feature space using explicit geometric correspondences. We further restrict cross-view retrieval to a geometry-guided local window to suppress spurious matches. Together, our method enables more robust refinement by grounding feature retrieval in the underlying 3D geometry.

\subsection{Geometric Correspondence Construction}
\label{sec:correspondence_construction}

Given a sparse training subset $\mathcal{V}_{\mathrm{tr}}$ from a posed multi-view capture, we optimize a 3DGS representation that evaluates synthesis at a novel viewpoint $t \notin \mathcal{V}_{\mathrm{tr}}$. At any intermediate stage, 3DGS produces an artifact-corrupted rendering $\tilde{I}^{t}$. Following the \emph{progressive refinement} paradigm~\cite{wu2025difix3d+}, we select the nearest training view $r \in \mathcal{V}_{\mathrm{tr}}$ as a reference to provide reliable scene information. To establish a physically-grounded information flow from the reference, we first obtain a metric depth map $D^{r}$ from an offline multi-view stereo pipeline~\cite{cao2024mvsformer++, lin2025depth}, providing scale-consistent geometry.

Let $\mathbf{u}_r \in \mathbb{R}^2$ and $\mathbf{u}_t \in \mathbb{R}^2$ denote pixel coordinates in the reference and target views, $\mathbf{K} \in \mathbb{R}^{3\times3}$ and $\mathbf{T} \in \mathbb{R}^{4\times4}$ denote the camera intrinsic and extrinsic, respectively. We establish the correspondence by projecting the 3D points, derived from $D^{r}$, onto the target image plane. Specifically, a reference point $\mathbf{x}$ is unprojected as $\mathbf{x} = \pi^{-1}(\mathbf{u}_r, D^{r}(\mathbf{u}_r), \mathbf{K}_r)$ and then projected to the target coordinate:
\begin{equation}
\mathbf{u}_t = \pi\!\left(\mathbf{K}_t\,\mathbf{T}_{t}\mathbf{T}_{r}^{-1}\,\mathbf{x}\right),
\end{equation}
where $\pi$ and $\pi^{-1}$ represent perspective projection and back-projection.

By forward-splatting \cite{niklaus2020softmaxsplatting} the reference coordinate map alongside an all-ones mask into the target camera, we directly obtain a dense geometric correspondence field $\mathcal{C}_{t\rightarrow r} \in \mathbb{R}^{H \times W \times 2}$ and a binary validity mask $M_{t\rightarrow r} \in \{0, 1\}^{H \times W}$. Here, $\mathcal{C}_{t\rightarrow r}$ acts as a spatial index mapping each target pixel $\mathbf{u}_t$ to its homologous reference counterpart, while $M_{t\rightarrow r}$ captures pixel visibility. Our objective is to learn a geometry-guided diffusion model $f_{\theta}$ that recovers the high-fidelity view: $\hat{I}^{t} = f_{\theta}(\tilde{I}^{t}\,;\, I^{r},\, \mathcal{C}_{t\rightarrow r},\, M_{t\rightarrow r})$.

\subsection{Geometry-Guided Cross-View Attention}
\label{sec:gca}

Recent reference-conditioned diffusion models for sparse-view synthesis~\cite{wu2025difix3d+,hirschorn2025splatent} typically leverage multi-view self-attention to aggregate features from the target view $F^t \in \mathbb{R}^{H/l \times W/l \times d}$ and reference view $F^r \in \mathbb{R}^{H/l \times W/l \times d}$, where $l$ is the spatial downsampling factor within the UNet blocks. However, during rendering refinement, queries derived from the artifact-prone target features $F^t$ suffer from query contamination. These corrupted signals mislead the feature retrieval process, subsequently propagating hallucinations in outputs. To rectify this, we introduce the \emph{Geometry-Guided Cross-View Attention (GCA)} module. The core mechanism is to bypass contaminated features by retrieving \emph{Geometry-Indexed Proxy Features} $F^{r \rightarrow t}$ directly from the clean reference feature space. We downsample $\mathcal{C}_{t\rightarrow r}$ and \(M_{t\to r}\) to match the feature resolution of each UNet block. The proxy features are formed by sampling homologous tokens:
\begin{equation}
F^{r\rightarrow t}(\mathbf{u}_t) = M_{t\rightarrow r}(\mathbf{u}_t) \odot \mathrm{Sample}\!\left(F^{r},\,\mathcal{C}_{t\rightarrow r}(\mathbf{u}_t)\right),
\end{equation}

\input{figure/figure_artifacts_removal}

where $\mathrm{Sample}(\cdot)$ denotes bilinear interpolation. Adopting $F^{r \rightarrow t}$ as the query $Q$ ensures that the attention is guided by clean reference content, effectively preventing the propagation of artifacts from the corrupted target into the output.

To further constrain retrieval around the geometric correspondence, we restrict attention to a local \(k\times k\) neighborhood \(\Omega\) centered at \(\mathcal{C}_{t\rightarrow r}(\mathbf{u}_t)\). For each target location \(\mathbf{u}_t\), we sample key and value features from the reference feature map at offsets \(\Delta \in \Omega\):

\begin{equation}
\begin{aligned}
K_{\Delta}(\mathbf{u}_t) &= \mathrm{Sample}\!\left(W_K F^r, \, \mathcal{C}_{t\rightarrow r}(\mathbf{u}_t) + \Delta\right), \\
V_{\Delta}(\mathbf{u}_t) &= \mathrm{Sample}\!\left(W_V F^r, \, \mathcal{C}_{t\rightarrow r}(\mathbf{u}_t) + \Delta\right),
\end{aligned}
\end{equation}

where \(W_K, W_V\) are linear projection matrices. The geometry-guided feature is then computed as

\begin{equation}
F_{\mathrm{geo}}^{t}(\mathbf{u}_t) = \sum_{\Delta\in\Omega} \mathrm{Softmax}_{\Delta}\!\left( \frac{\langle Q(\mathbf{u}_t), K_{\Delta}(\mathbf{u}_t)\rangle}{\sqrt{d}} \right) V_{\Delta}(\mathbf{u}_t).
\end{equation}

This local constraint reduces spurious long-range matches and encourages geometrically consistent retrieval. The geometry-guided feature \(F^{t}_{\mathrm{geo}}\) is then integrated into the backbone through an adaptive fusion mechanism. Specifically, we predict a spatial gating map $w = \sigma(\mathrm{MLP}([F^{t}, F^{t}_{\mathrm{geo}}]))$, which controls the contribution of the GCA branch:
\begin{equation}
F^{t}(\mathbf{u}_t) \leftarrow (1-w(\mathbf{u}_t)) \odot F^{t}(\mathbf{u}_t) + w(\mathbf{u}_t) \odot F^{t}_{\mathrm{geo}}(\mathbf{u}_t).
\end{equation}

This allows geometry-guided features to be injected without sacrificing global semantic context, helping maintain robustness when correspondences are weak.

\subsection{Training Objective}
\label{sec:training_objective}
We supervise the refined output $\hat{I}$ against the ground-truth $I$ using a combination of reconstruction, perceptual, and style losses. Specifically, we use a pixel-wise $\ell_2$ reconstruction loss $\mathcal{L}_{\text{recon}}$, a perceptual loss $\mathcal{L}_{\text{lpips}}$, and a style loss $\mathcal{L}_{\text{gram}}$ \cite{wu2025difix3d+, reda2022film} to encourage sharper textures:
\begin{equation}
    \mathcal{L}_{\text{gram}} = \frac{1}{L} \sum_{l=1}^{L} \beta_l \|G_l(\hat{I}) - G_l(I)\|_2,
\end{equation}
\input{tables/table_artifacts_removal}
where $G_l(I) = \phi_l(I)^\top \phi_l(I)$ denotes the Gram matrix of VGG-16 features $\phi_l(\cdot)$ at layer $l$, and $\beta_l$ represents the layer-specific weight. The full objective is
\begin{equation}
    \mathcal{L} = \lambda_{\text{recon}} \mathcal{L}_{\text{recon}} + \lambda_{\text{lpips}} \mathcal{L}_{\text{lpips}} + \lambda_{\text{gram}} \mathcal{L}_{\text{gram}},
\end{equation}
where $\lambda_{\text{recon}}$, $\lambda_{\text{lpips}}$, and $\lambda_{\text{gram}}$ denote the weight coefficients.

\paragraph{Discussion.}
GeoQuery offers two practical advantages. First, when reference projections are invalid or occluded, the validity mask disables local retrieval, and the adaptive fusion falls back to the global branch for semantic completion. This helps maintain stable refinement when geometric correspondences are weak or missing. Second, the local \(k\times k\) window improves both accuracy and efficiency. It reduces spurious matches, as supported by Fig.~\ref{fig:ablation_window_size}, and lowers the complexity of cross-view attention from \(\mathcal{O}(N^2)\) to \(\mathcal{O}(Nk^2)\). This linear scaling makes the module more practical for high-resolution inputs by alleviating the computational burden of global attention.

%% file: figure/figure_artifacts_removal.tex
\begin{figure*}[t]

  \centering

  \includegraphics[width=\textwidth]{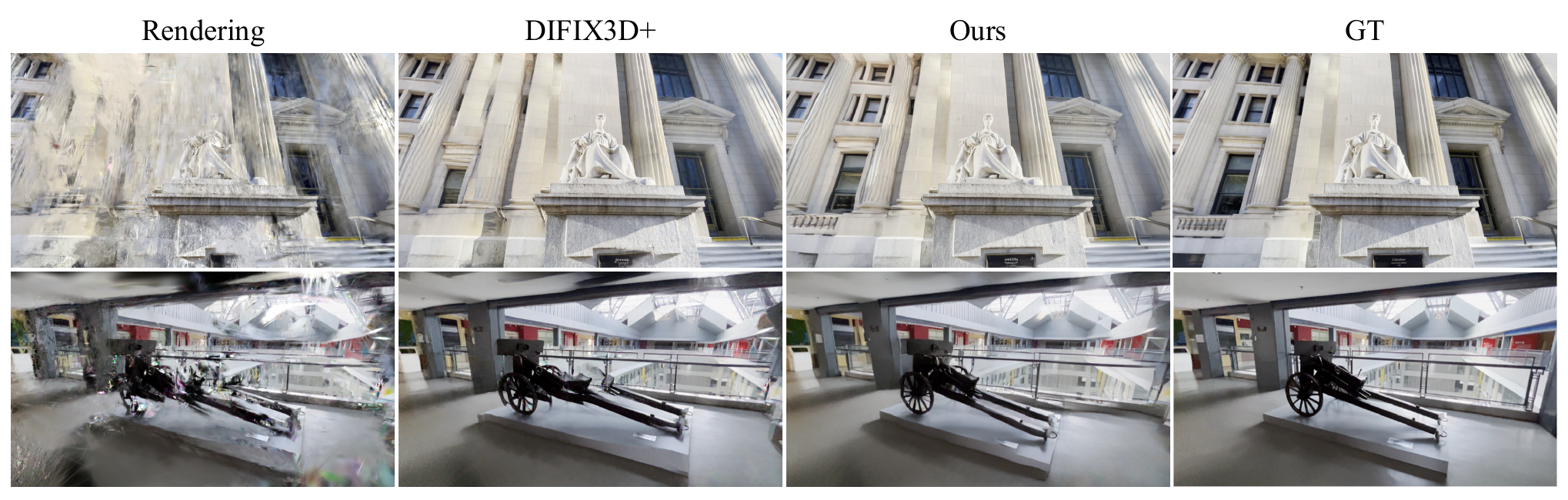}

  \caption{Qualitative comparisons on artifact removal. From left to right: the artifact-corrupted 3DGS rendering, DIFIX3D+~\cite{wu2025difix3d+}, our GeoQuery, and the ground truth. The examples illustrate that GeoQuery can more reliably restore rendering structure, producing results that are visually closer to the ground truth.}

  \label{fig:artifacts-removal}

\end{figure*}

%% file: tables/table_artifacts_removal.tex
\label{sec:artifacts-removal}
\begin{table}
\centering
\small
\setlength{\tabcolsep}{3pt}
\renewcommand{\arraystretch}{0.95}
\caption{Quantitative comparison for artifact removal on the DL3DV-Benchmark test set. GeoQuery achieves the best overall performance across all metrics.}
\label{tab:artifacts-removal}
\begin{tabular}{lcccc}
\toprule
\textbf{Method} & \textbf{PSNR}$\uparrow$ & \textbf{SSIM}$\uparrow$ & \textbf{LPIPS}$\downarrow$ & \textbf{FID}$\downarrow$ \\
\midrule
DIFIX3D+(w/o. ref) \cite{wu2025difix3d+}           & 18.26 & 0.493 & 0.388 & 21.04 \\
DIFIX3D+ \cite{wu2025difix3d+}                    & 18.79 & 0.529 & 0.348 & 12.83 \\
GeoQuery                             & \textbf{19.88} & \textbf{0.566} & \textbf{0.314} & \textbf{10.20} \\
\bottomrule
\end{tabular}
\end{table}

%% file: main/05_experiment.tex
\section{Experiment}

\subsection{Experimental Setup}
\label{sec:setup}
\input{main/07_figures_pages}
\paragraph{Datasets.}
We train GeoQuery on the DL3DV-Benchmark~\cite{ling2024dl3dv} using 123 scenes to construct $\sim$100K training pairs. To simulate corrupted inputs, we optimize 3DGS on subsampled camera trajectories and render novel viewpoints as artifact-prone inputs. These are paired with original captured images for supervision, while the subsampled frames provide reference guidance. Training-view metric depth is precomputed via MVSFormer++~\cite{cao2024mvsformer++}. Evaluation is conducted on 12 DL3DV-Benchmark scenes and the Mip-NeRF360 dataset~\cite{barron2022mip}, with the latter following the split protocol established in ReconFusion~\cite{wu2024reconfusion}.

\paragraph{Implementation Details.}
Our framework is built on the pre-trained SD-Turbo~\cite{sauer2024adversarial}. The Geometry-Guided Cross-View Attention (GCA) module is initialized from pre-trained self-attention weights and integrated into the low-resolution UNet blocks. We train at $576\times1024$ resolution for 100K iterations using the AdamW optimizer (learning rate $2\times10^{-5}$, batch size 1) on a single NVIDIA A100 GPU. For 3DGS reconstruction, Depth Anything v3~\cite{lin2025depth} provides the necessary geometric guidance.

\input{tables/table_sparse_nvs}
\input{tables/table_qc_region_study}
\input{figure/figure_ablation_window_size}

\subsection{3DGS Artifacts Removal}

We evaluate GeoQuery on 12 DL3DV-Benchmark scenes for artifact removal. As shown in Table~\ref{tab:artifacts-removal}, GeoQuery consistently outperforms DIFIX3D+~\cite{wu2025difix3d+} across all metrics, notably achieving a 1.09 dB gain in PSNR and a 2.63 reduction in FID. These results support the effectiveness of our geometry-guided approach in promoting more consistent refinement. Qualitative results in Fig.~\ref{fig:artifacts-removal} and Fig.~\ref{fig:artifacts-removal-figure-only} confirm its robustness: while baselines often propagate incorrect textures into noisy regions, GeoQuery recovers accurate details.
\input{tables/table_ablation_artifacts_removal}
\input{tables/table_ablation_nvs}

\subsection{Region-level Study on Query Contamination}
\label{sec:qc_study}
We further analyze query contamination on the DL3DV dataset by partitioning each 3DGS rendering according to its pixel-wise error \(e(\mathbf{u})\) with respect to the ground truth. Using a threshold \(\tau=30\), we define a low-error region \(\{\mathbf{u}\mid e(\mathbf{u}) \le \tau\}\) and a high-error region \(\{\mathbf{u}\mid e(\mathbf{u}) > \tau\}\). As shown in Table~\ref{tab:qc_analysis}, DIFIX3D+ degrades the low-error region while providing only limited recovery in the high-error region. This behavior is consistent with our observation that artifact-prone queries retrieve mismatched content from the reference view, which inherently damages refinement. In contrast, by addressing query contamination, GeoQuery achieves better refinement in both regions.

\subsection{Sparse-View Reconstruction}
\label{sec:sparse_view}
\input{figure/figure_nvs_360_dl3dv}
\input{tables/table_ablation_epipolar_attention}
\input{figure/figure_ablation_gca}
We evaluate GeoQuery on the DL3DV-Benchmark~\cite{ling2024dl3dv} and Mip-NeRF360~\cite{barron2022mip} datasets under extreme sparsity (3, 6, and 9 views). As shown in Table~\ref{tab:quant_main}, GeoQuery consistently achieves better PSNR and SSIM than baseline methods, with the largest gains in the challenging 3-view regime. Qualitative results in Fig.~\ref{fig:360-dl3dv} show that GeoQuery produces cleaner and more plausible renderings than baseline methods. For qualitative comparison, both GeoQuery and DIFIX3D+ in Fig.~\ref{fig:360-dl3dv} are shown after their optional post-processing refinement. Fig.~\ref{fig:same-scene-ablation} further shows that GeoQuery degrades more gracefully as input-view sparsity increases. On a single NVIDIA A100 at 1237$\times$822 resolution, our implementation uses 21.13\,GB peak memory and takes $\sim$1.2\,s per image for diffusion-based refinement.

\subsection{Ablation Study}
\paragraph{Effectiveness of GCA and Proxy Queries.}
We validate the GCA module across both artifact removal (Table~\ref{tab:ablation_2d_restoration}) and sparse-view view synthesis (Table~\ref{tab:ablation}). Quantitative comparisons consistently demonstrate that substituting rendering-based queries with proxy queries yields superior fidelity, confirming the critical role of clean geometric guidance in mitigating query contamination. As visualized in Fig.~\ref{fig:ablation_anchor_attn}, our approach effectively mitigates the inconsistent refinements caused by mismatched cross-view retrieval. We further compare GCA with standard epipolar attention. As shown in Table~\ref{tab:ablation_epi}, introducing proxy query features already improves over epipolar attention, while replacing epipolar attention with GCA yields further gains across all metrics.

\paragraph{Ablation of Different Window Sizes.}
We analyze the effect of the attention window size  $k$ in Geometry-Guided Cross-View Attention on our 3DGS artifact removal test set (Sec.~\ref{sec:artifacts-removal}); Fig.~\ref{fig:ablation_window_size} reports the FID results.
A moderate window ($k{=}3$) achieves the lowest FID.
Both smaller windows and larger windows degrade performance, suggesting a trade-off between insufficient local evidence and increased matching ambiguity.
We also include an unconstrained cross-attention variant (``cross-attn'', i.e., GCA without window restriction) as a baseline, which performs worse than the best windowed setting.

%% file: main/07_figures_pages.tex
\begin{figure*}
  \centering
  \includegraphics[width=\textwidth]{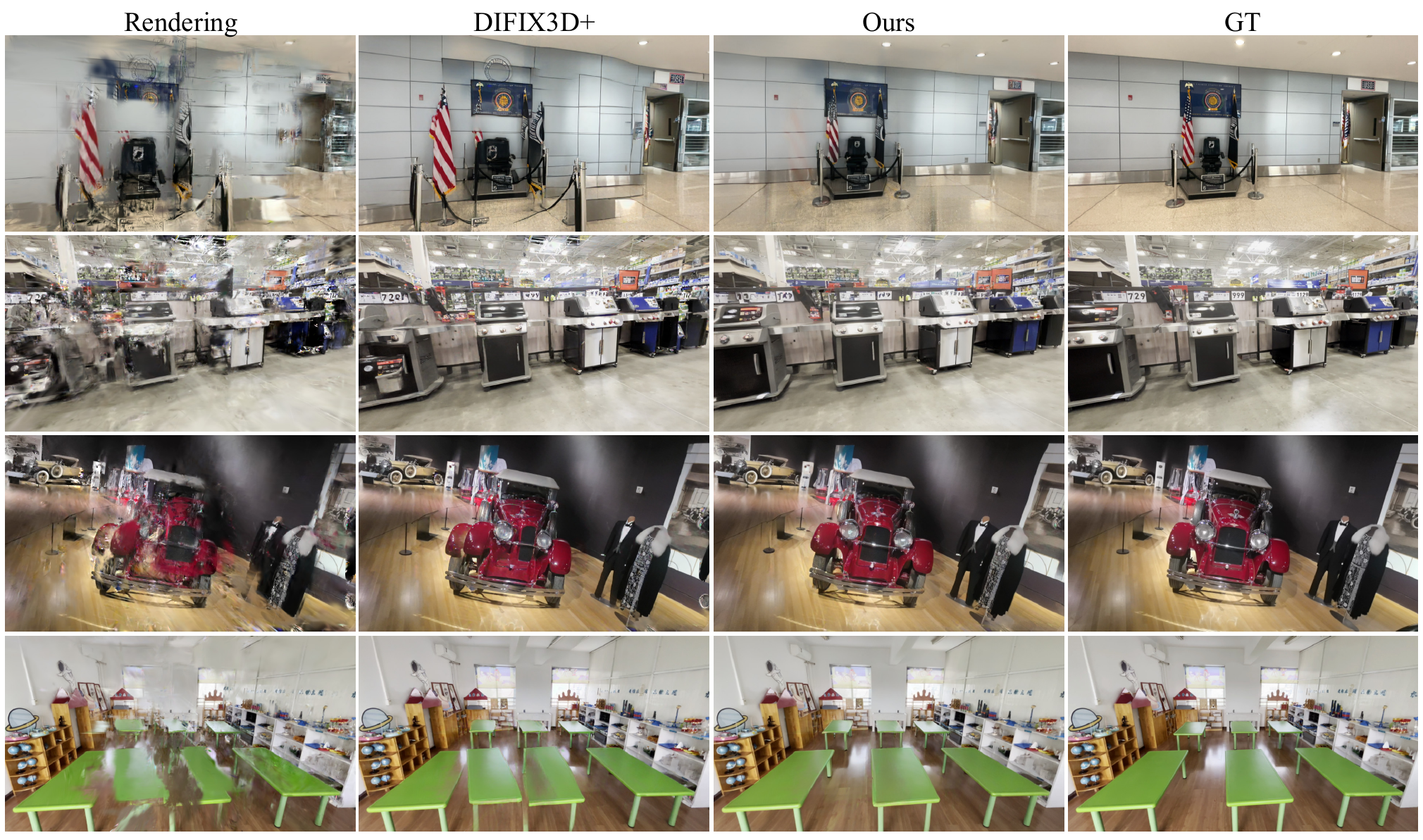}
  \caption{More Qualitative comparisons on artifact removal.}
  \label{fig:artifacts-removal-figure-only}
\end{figure*}

\begin{figure*}
  \centering
  \includegraphics[width=\textwidth]{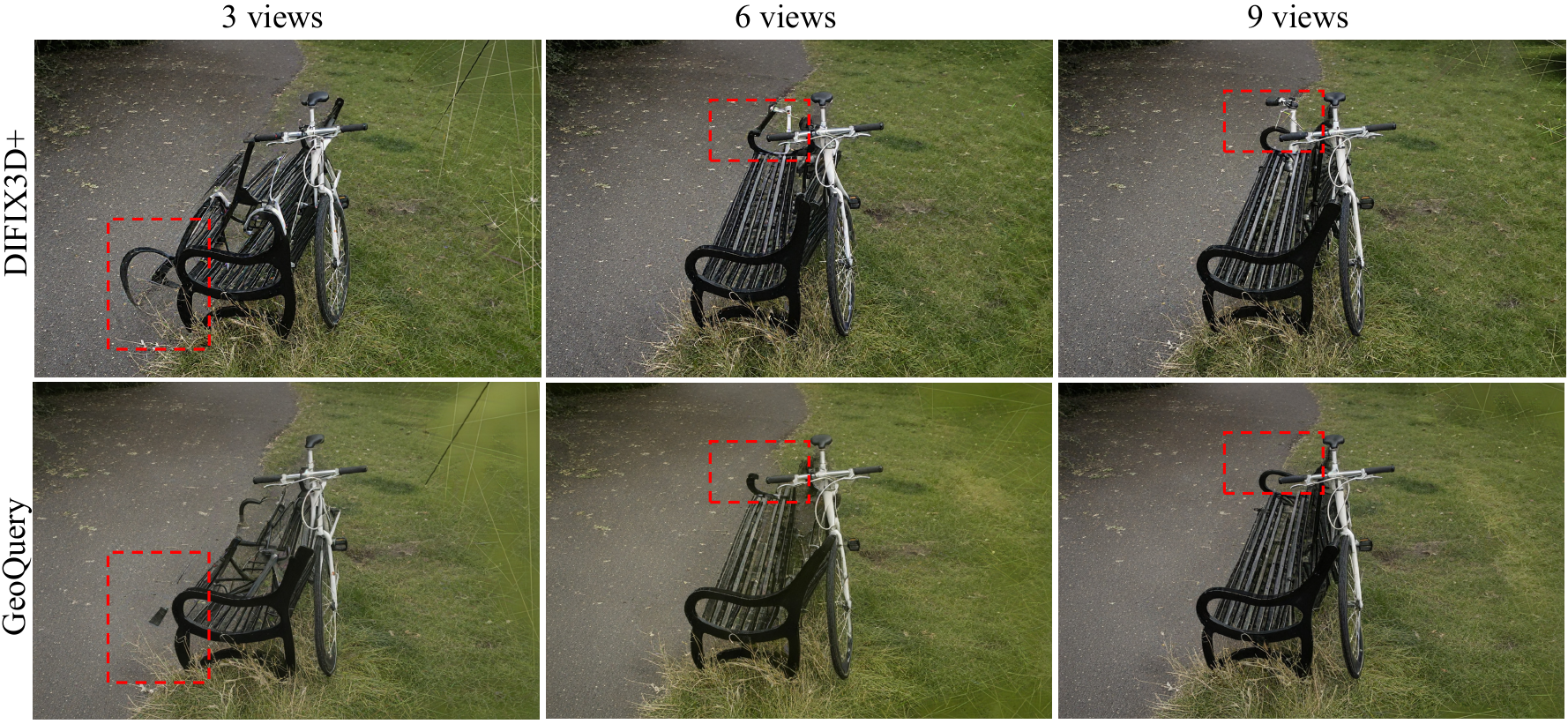}
  \caption{Same-scene comparison under varying input views on the \textit{bicycle} scene of Mip-NeRF360 \cite{barron2022mip}. We fix the same target novel view and compare DIFIX3D+ and GeoQuery under 3, 6, and 9 training views. GeoQuery maintains more stable reconstruction quality as view sparsity increases.}
  \label{fig:same-scene-ablation}
\end{figure*}

\begin{figure*}
  \centering
  \includegraphics[width=\textwidth]{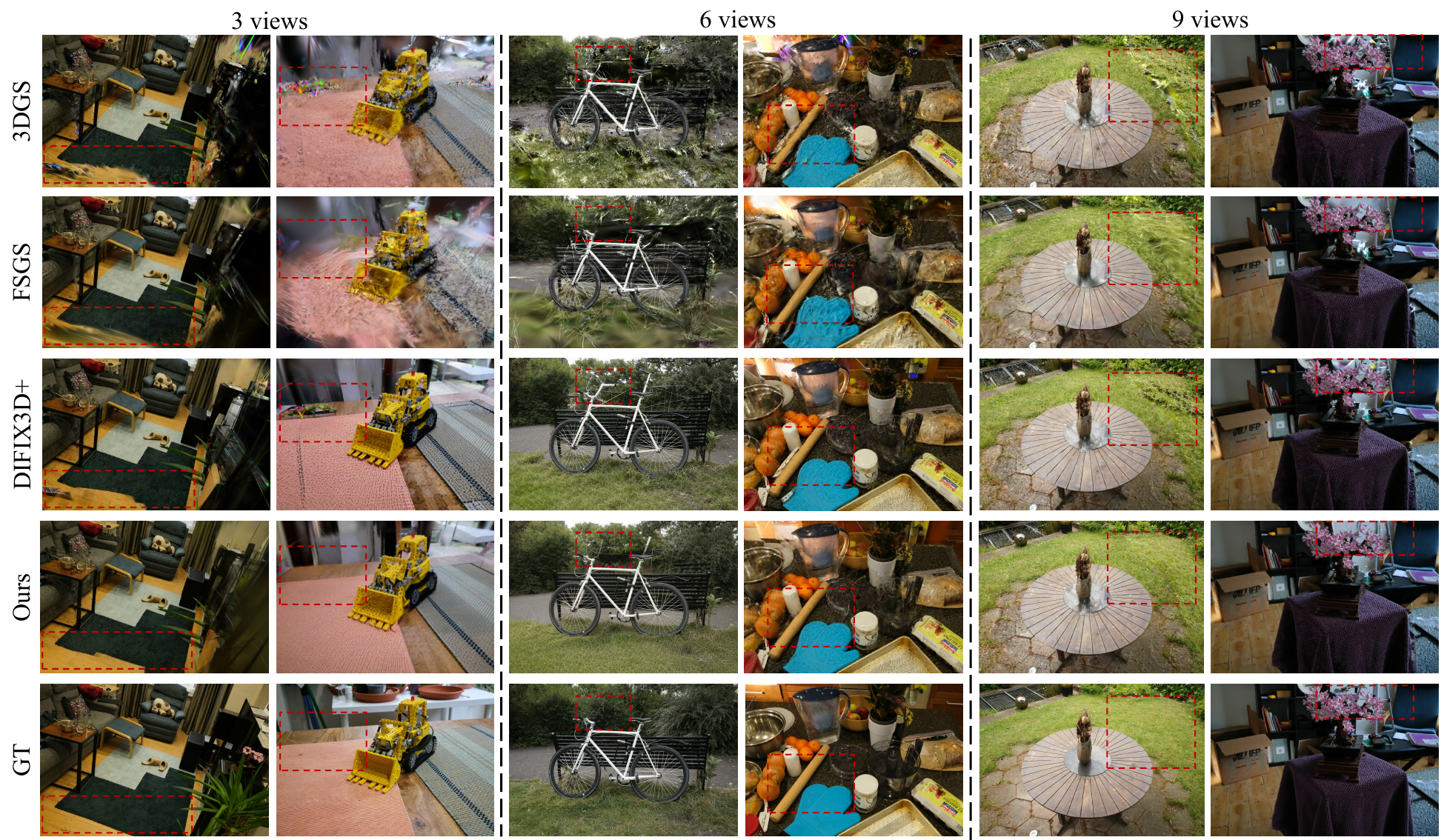}
  \caption{More Qualitative results on Mip-NeRF360 dataset \cite{barron2022mip}.}
  \label{fig:sparse-nvs-360}
\end{figure*}

\begin{figure*}
  \centering
  \includegraphics[width=\textwidth]{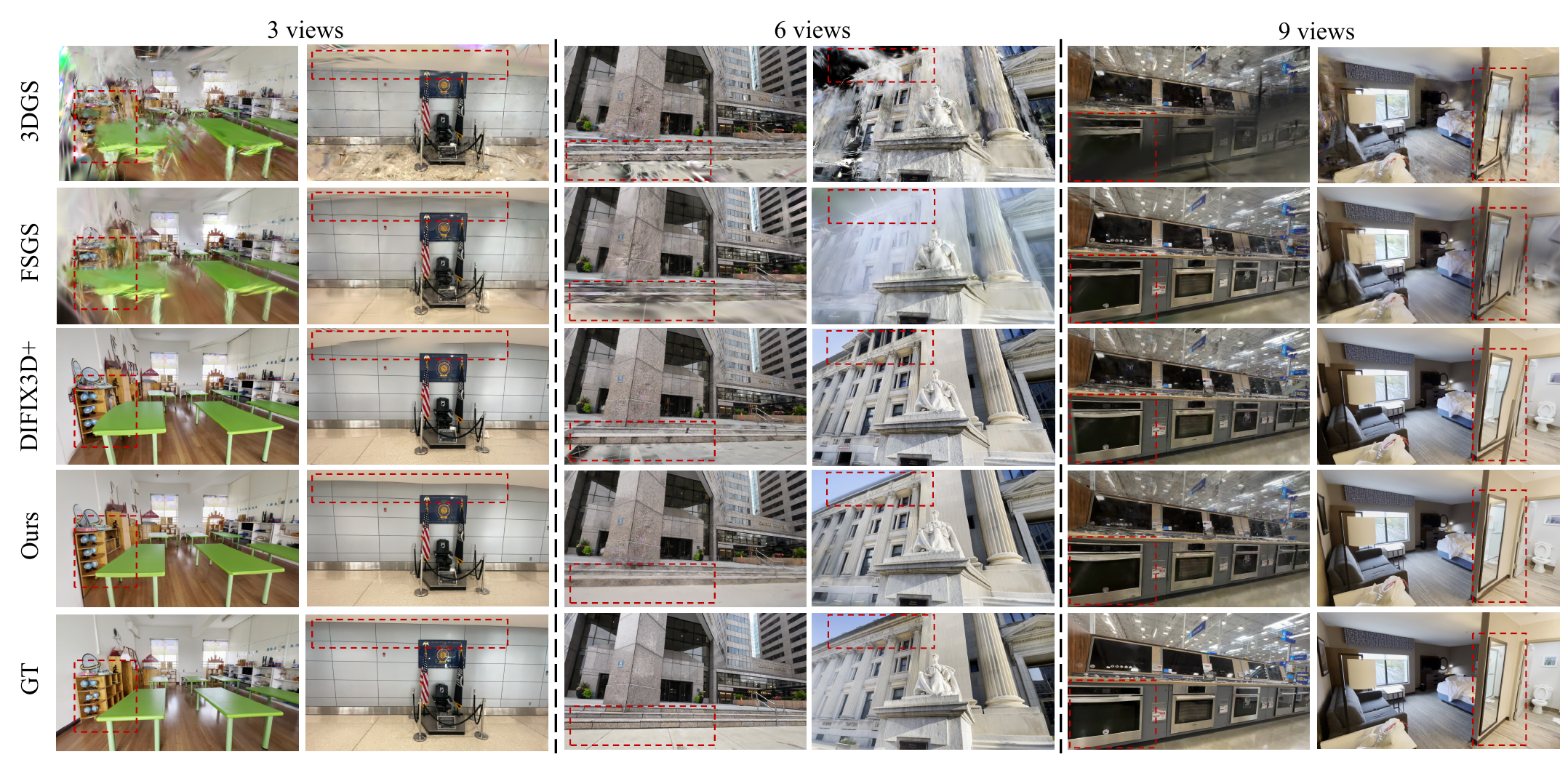}
  \caption{More Qualitative results on DL3DV-Benchmark dataset \cite{ling2024dl3dv}.}
  \label{fig:sparse-nvs-dl3dv}
\end{figure*}

%% file: tables/table_sparse_nvs.tex
\begin{table*}[t]
\centering
\caption{Quantitative comparison of rendering quality on Mip-NeRF360~\cite{barron2022mip} and DL3DV-Benchmark~\cite{ling2024dl3dv} under different numbers of input views (3, 6, and 9 views).
$^{\ddagger}$ denotes results reproduced using the official implementations of the corresponding methods.}
\label{tab:quant_main}

\setlength{\tabcolsep}{4pt}
\renewcommand{\arraystretch}{0.95}

\begin{tabular}{lcccccccccccc}
\toprule
& \multicolumn{4}{c}{PSNR$\uparrow$}
& \multicolumn{4}{c}{SSIM$\uparrow$}
& \multicolumn{4}{c}{LPIPS$\downarrow$} \\
\cmidrule(lr){2-5}\cmidrule(lr){6-9}\cmidrule(lr){10-13}
Method
& 3-view & 6-view & 9-view & Avg.
& 3-view & 6-view & 9-view & Avg.
& 3-view & 6-view & 9-view & Avg. \\
\midrule

\multicolumn{13}{l}{\textbf{Mip-NeRF360~\cite{barron2022mip}}} \\
\midrule
3DGS$^{\ddagger}$~\cite{kerbl20233d}
& 13.06 & 14.96 & 16.79 & 14.94
& 0.251 & 0.355 & 0.447 & 0.351
& 0.576 & 0.505 & 0.446 & 0.509 \\
FSGS$^{\ddagger}$~\cite{zhu2024fsgs}
& 13.98 & 15.92 & \sbest 17.74 & 15.88
& \sbest 0.310 & \sbest 0.409 & \best 0.488 & \sbest 0.403
& 0.575 & 0.513 & 0.464 & 0.517 \\
DIFIX3D$^{\ddagger}$~\cite{wu2025difix3d+}
& \sbest 14.15 & \sbest 16.14 & 17.54 & \sbest 15.94
& 0.29 & 0.378 & 0.445 & 0.371
& \best 0.522 & \best 0.422 & \sbest 0.356 & \best 0.433 \\
GeoQuery (Ours)
& \best 15.07 & \best 16.93 & \best 18.22 & \best 16.74
& \best 0.334 & \best 0.411 & \sbest 0.468 & \best 0.404
& \sbest 0.531 & \sbest 0.423 & \best 0.355 & \sbest 0.436 \\
\midrule

\multicolumn{13}{l}{\textbf{DL3DV-Benchmark~\cite{ling2024dl3dv}}} \\
\midrule
3DGS$^{\ddagger}$~\cite{kerbl20233d}
& 13.89 & 16.68 & 18.63 & 16.40
& 0.502 & 0.607 & 0.685 & 0.598
& 0.543 & 0.412 & 0.323 & 0.426 \\
FSGS$^{\ddagger}$~\cite{zhu2024fsgs}
& \sbest 15.22 & 17.83 & \sbest 20.15 & 17.73
& \sbest 0.602 & \sbest 0.690 & \best 0.752 & \best 0.681
& 0.466 & 0.371 & 0.307 & 0.381 \\
DIFIX3D$^{\ddagger}$~\cite{wu2025difix3d+}
& 15.20 & \sbest 18.06 & 19.97 & \sbest 17.74
& 0.579 & 0.670 & 0.728 & 0.659
& \sbest 0.447 & \sbest 0.316 & \sbest 0.251 & \sbest 0.338 \\
GeoQuery (Ours)
& \best 15.98 & \best 18.60 & \best 20.20 & \best 18.26
& \best 0.614 & \best 0.692 & \sbest 0.738 & \best 0.681
& \best 0.441 & \best 0.313 & \best 0.249 & \best 0.334 \\
\bottomrule
\end{tabular}
\end{table*}

%% file: tables/table_qc_region_study.tex
\begin{table}
\centering
\small
\caption{Region-level PSNR analysis under error-threshold partition. Average PSNR over low-error \((e(\mathbf{u}) \le \tau)\) and high-error \((e(\mathbf{u}) > \tau)\) regions, with \(\tau=30\).}
\label{tab:qc_analysis}
\begin{tabular}{lccc}
\toprule
\textbf{Region} & \textbf{3DGS} & \textbf{DIFIX3D+} & \textbf{GeoQuery (Ours)} \\
\midrule
Low-error (\(e(\mathbf{u}) \le \tau\)) & 25.82 & 25.07 \textcolor{red}{\small(-0.75)} & \textbf{26.19} \textcolor{blue}{\small(+0.37)} \\
High-error (\(e(\mathbf{u}) > \tau\)) & 11.16 & 13.16 \textcolor{blue}{\small(+2.00)} & \textbf{15.19} \textcolor{blue}{\small(+4.03)} \\
\bottomrule
\end{tabular}
\end{table}

%% file: figure/figure_ablation_window_size.tex
\begin{figure}
  \centering
  \includegraphics[width=0.6\columnwidth]{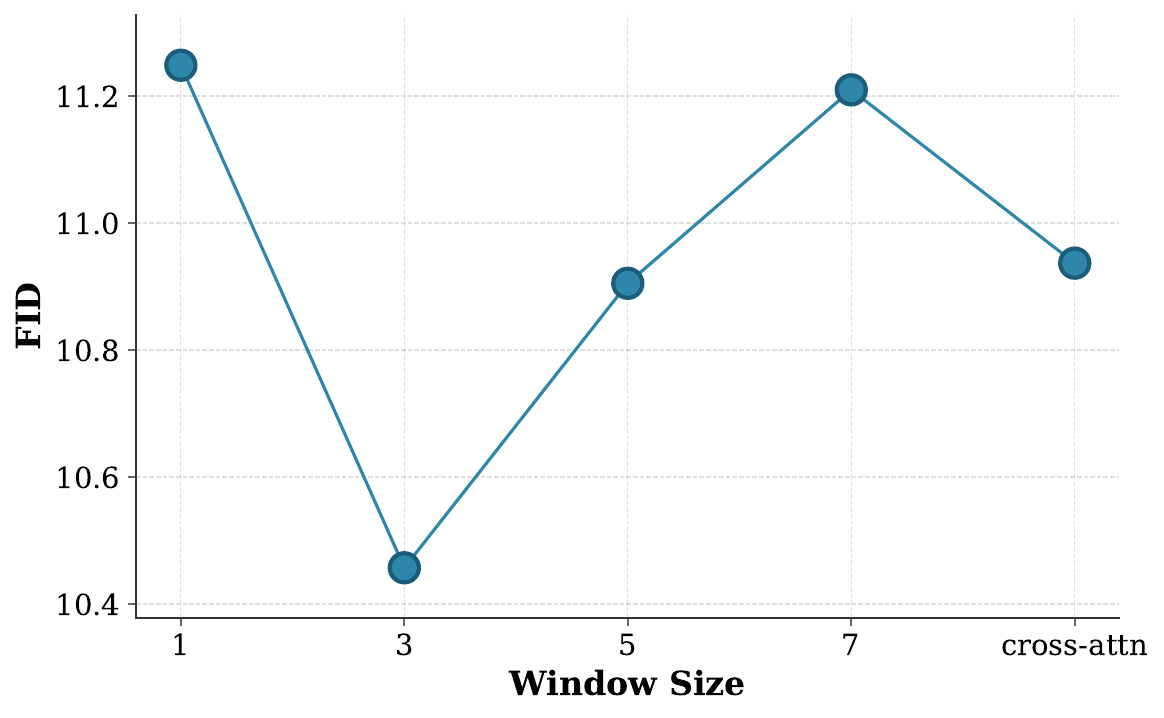}
  \caption{Ablation on window size in GCA. We report FID on the rendering artifact removal task when varying the local attention window size $k$.}
  \label{fig:ablation_window_size}
\end{figure}

%% file: tables/table_ablation_artifacts_removal.tex
\begin{table}
\centering
\caption{Ablation of individual components for 2D artifact removal. SA: Multi-view Self-Attention; GCA/R: GCA with rendering query; GCA/P: GCA with proxy query; AF: Adaptive Feature Fusion gating mechanism.}
\label{tab:ablation_2d_restoration}
\small
\setlength{\tabcolsep}{4pt}
\renewcommand{\arraystretch}{0.9}
\begin{tabular}{cccccccc}
\toprule
\textbf{SA} & \textbf{GCA/R} & \textbf{GCA/P} & \textbf{AF} & \textbf{PSNR}$\uparrow$ & \textbf{SSIM}$\uparrow$ & \textbf{LPIPS}$\downarrow$ & \textbf{FID}$\downarrow$ \\
\midrule
\checkmark &            &            &            & 18.79 & 0.529 & 0.348 & 12.83 \\
\checkmark & \checkmark &            & \checkmark & 19.42 & 0.549 & 0.332 & 11.60 \\
\checkmark &            & \checkmark &            & 19.57 & 0.556 & 0.322 & 11.11 \\
\checkmark &            & \checkmark & \checkmark & \textbf{19.73} & \textbf{0.561} & \textbf{0.319} & \textbf{10.90} \\
\bottomrule
\end{tabular}
\end{table}

%% file: tables/table_ablation_nvs.tex
\begin{table}
\centering
\small
\setlength{\tabcolsep}{6pt}
\renewcommand{\arraystretch}{0.9}
\caption{Ablation study of our method on Mip-NeRF360~\cite{barron2022mip} under the 3-view setting.}
\label{tab:ablation}
\begin{tabular}{lccc}
\toprule
\textbf{Method} & \textbf{PSNR}$\uparrow$ & \textbf{SSIM}$\uparrow$ & \textbf{LPIPS}$\downarrow$ \\
\midrule
(A) Global attention only      & 14.57 & 0.319 & 0.562 \\
(B) + GCA ($Q=\text{Rendering}$) & 14.81 & 0.326 & 0.532 \\
(C) + GCA ($Q=\text{Proxy}$) & \textbf{15.07} & \textbf{0.334} & \textbf{0.531} \\
\bottomrule
\end{tabular}
\end{table}

%% file: figure/figure_nvs_360_dl3dv.tex
\begin{figure*}[t]
  \centering
  \includegraphics[width=\textwidth]{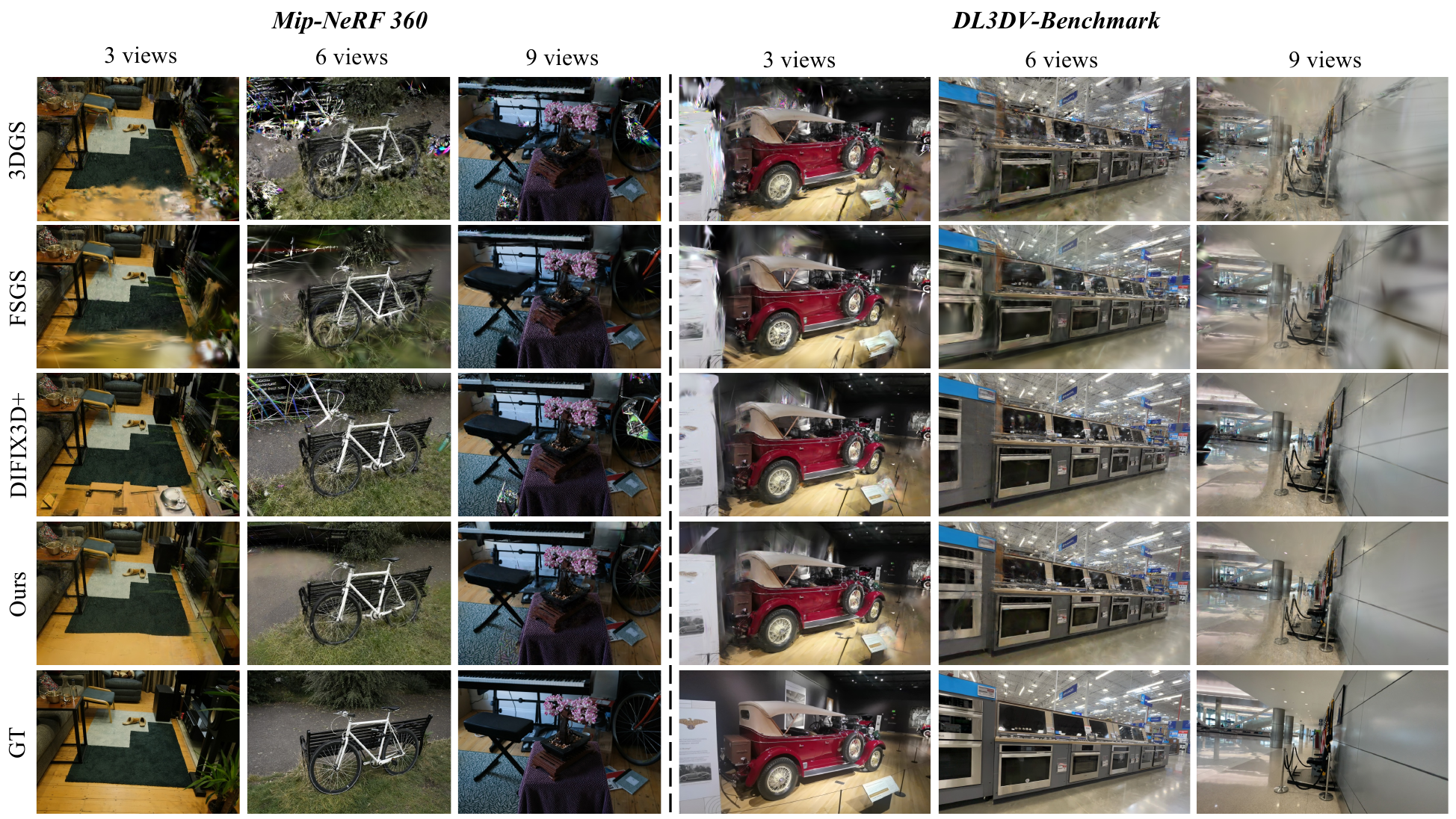}
  \caption{Visual comparisons on Mip-NeRF360 dataset and DL3DV dataset between our GeoQuery and baseline methods, including 3DGS~\cite{kerbl20233d}, FSGS~\cite{zhu2024fsgs}, and DIFIX3D+~\cite{wu2025difix3d+}. 
Compared to the baselines, our method produces more reliable renderings.}
  \label{fig:360-dl3dv}
  \vspace{-0.3cm}
\end{figure*}

%% file: tables/table_ablation_epipolar_attention.tex
\begin{table}
\centering
\small
\caption{Comparison with epipolar attention. Quantitative comparison on the artifact removal task.}

\label{tab:ablation_epi}
\begin{tabular}{lcccc}
\toprule
\textbf{Method} & \textbf{PSNR}$\uparrow$ & \textbf{SSIM}$\uparrow$ & \textbf{LPIPS}$\downarrow$ & \textbf{FID}$\downarrow$ \\
\midrule
GeoQuery (Epi.Attn)         & 19.21 & 0.542 & 0.338 & 12.16 \\
GeoQuery (Proxy+Epi.Attn)   & 19.73 & 0.554 & 0.322 & 11.05 \\
GeoQuery (Proxy+GCA)        & \textbf{19.88} & \textbf{0.566} & \textbf{0.314} & \textbf{10.20} \\
\bottomrule
\end{tabular}
\end{table}

%% file: figure/figure_ablation_gca.tex
\begin{figure}
  \centering
  \includegraphics[width=\columnwidth]{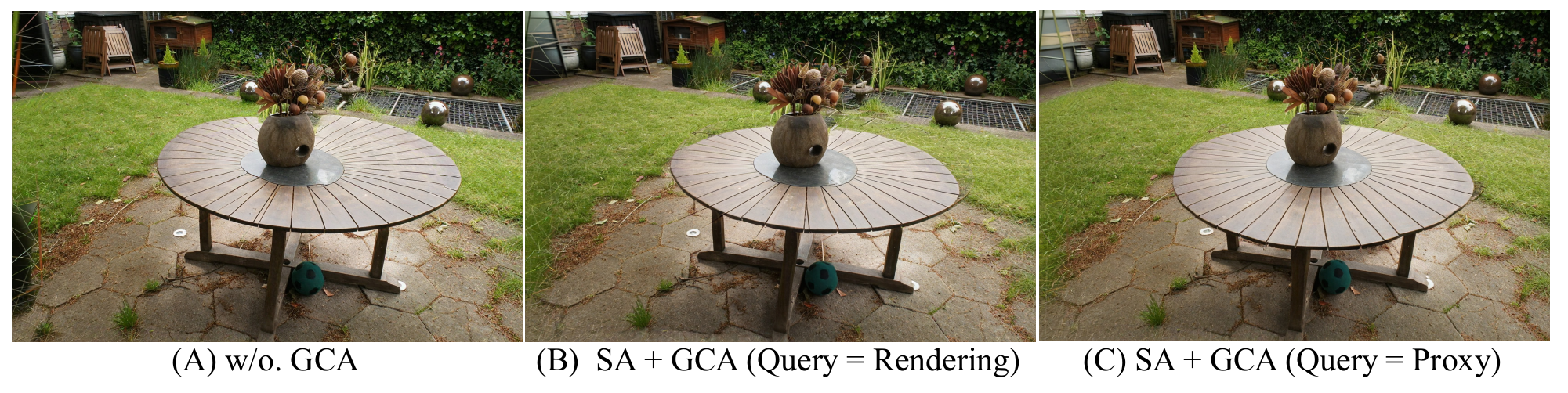}
  \caption{Qualitative ablation for GCA effects. From left to right - (A) Multi-view Self-Attention only. (B) Self-Attention + GCA (query = rendering). (C) Self-Attention + GCA (query = proxy)}
  \label{fig:ablation_anchor_attn}
\end{figure}

%% file: main/06_conclusion.tex
\section{Limitations and Conclusion}
\label{sec:conclusion}
\paragraph{Limitations and future works.} While GeoQuery effectively improves refinement via geometric guidance, its reliance on explicit correspondences introduces limitations in textureless or specular regions where depth estimation typically fails. Moreover, when correspondences are absent due to extreme viewpoint disparities, the restoration quality depends solely on the generative capacity of the diffusion model. For future works, a promising direction is to leverage more powerful diffusion models for refinement.
\paragraph{Conclusions.} We presented GeoQuery, a geometry-guided diffusion framework for sparse-view 3D Gaussian Splatting. To resolve the query contamination issue in cross-view attention mechanisms, we introduce Geometry-Indexed Proxy Features. By anchoring feature retrieval via Geometry-Guided Cross-View Attention, GeoQuery effectively suppresses inconsistent refinement. Extensive evaluations on the DL3DV-Benchmark and Mip-NeRF360 datasets demonstrate that our framework consistently outperforms baselines in both artifact removal and novel view synthesis.

%% file: sample-base.bib
@inproceedings{rombach2022ldm,
  title={High-resolution image synthesis with latent diffusion models},
  author={Rombach, Robin and Blattmann, Andreas and Lorenz, Dominik and Esser, Patrick and Ommer, Bj{\"o}rn},
  booktitle={Proceedings of the IEEE/CVF conference on computer vision and pattern recognition},
  pages={10684--10695},
  year={2022}
}

@article{ho2020ddpm,
  title={Denoising diffusion probabilistic models},
  author={Ho, Jonathan and Jain, Ajay and Abbeel, Pieter},
  journal={Advances in neural information processing systems},
  volume={33},
  pages={6840--6851},
  year={2020}
}

@article{song2020sde,
  title={Score-based generative modeling through stochastic differential equations},
  author={Song, Yang and Sohl-Dickstein, Jascha and Kingma, Diederik P and Kumar, Abhishek and Ermon, Stefano and Poole, Ben},
  journal={arXiv preprint arXiv:2011.13456},
  year={2020}
}

@article{mildenhall2021nerf,
  title={Nerf: Representing scenes as neural radiance fields for view synthesis},
  author={Mildenhall, Ben and Srinivasan, Pratul P and Tancik, Matthew and Barron, Jonathan T and Ramamoorthi, Ravi and Ng, Ren},
  journal={Communications of the ACM},
  volume={65},
  number={1},
  pages={99--106},
  year={2021},
  publisher={ACM New York, NY, USA}
}

@inproceedings{wu2025difix3d+,
  title={Difix3d+: Improving 3d reconstructions with single-step diffusion models},
  author={Wu, Jay Zhangjie and Zhang, Yuxuan and Turki, Haithem and Ren, Xuanchi and Gao, Jun and Shou, Mike Zheng and Fidler, Sanja and Gojcic, Zan and Ling, Huan},
  booktitle={Proceedings of the Computer Vision and Pattern Recognition Conference},
  pages={26024--26035},
  year={2025}
}

@inproceedings{wu2025genfusion,
  title={Genfusion: Closing the loop between reconstruction and generation via videos},
  author={Wu, Sibo and Xu, Congrong and Huang, Binbin and Geiger, Andreas and Chen, Anpei},
  booktitle={Proceedings of the Computer Vision and Pattern Recognition Conference},
  pages={6078--6088},
  year={2025}
}

@article{yin2025gsfixer,
  title={Gsfixer: Improving 3d gaussian splatting with reference-guided video diffusion priors},
  author={Yin, Xingyilang and Zhang, Qi and Chang, Jiahao and Feng, Ying and Fan, Qingnan and Yang, Xi and Pun, Chi-Man and Zhang, Huaqi and Cun, Xiaodong},
  journal={arXiv preprint arXiv:2508.09667},
  year={2025}
}

@article{cao2024mvsformer++,
  title={Mvsformer++: Revealing the devil in transformer's details for multi-view stereo},
  author={Cao, Chenjie and Ren, Xinlin and Fu, Yanwei},
  journal={arXiv preprint arXiv:2401.11673},
  year={2024}
}

@article{kerbl20233d,
  title={3D Gaussian splatting for real-time radiance field rendering.},
  author={Kerbl, Bernhard and Kopanas, Georgios and Leimk{\"u}hler, Thomas and Drettakis, George},
  journal={ACM Trans. Graph.},
  volume={42},
  number={4},
  pages={139--1},
  year={2023}
}

@inproceedings{zhu2024fsgs,
  title={Fsgs: Real-time few-shot view synthesis using gaussian splatting},
  author={Zhu, Zehao and Fan, Zhiwen and Jiang, Yifan and Wang, Zhangyang},
  booktitle={European conference on computer vision},
  pages={145--163},
  year={2024},
  organization={Springer}
}

@inproceedings{wu2024reconfusion,
  title={Reconfusion: 3d reconstruction with diffusion priors},
  author={Wu, Rundi and Mildenhall, Ben and Henzler, Philipp and Park, Keunhong and Gao, Ruiqi and Watson, Daniel and Srinivasan, Pratul P and Verbin, Dor and Barron, Jonathan T and Poole, Ben and others},
  booktitle={Proceedings of the IEEE/CVF conference on computer vision and pattern recognition},
  pages={21551--21561},
  year={2024}
}

@inproceedings{barron2022mip,
  title={Mip-nerf 360: Unbounded anti-aliased neural radiance fields},
  author={Barron, Jonathan T and Mildenhall, Ben and Verbin, Dor and Srinivasan, Pratul P and Hedman, Peter},
  booktitle={Proceedings of the IEEE/CVF conference on computer vision and pattern recognition},
  pages={5470--5479},
  year={2022}
}

@inproceedings{ling2024dl3dv,
  title={Dl3dv-10k: A large-scale scene dataset for deep learning-based 3d vision},
  author={Ling, Lu and Sheng, Yichen and Tu, Zhi and Zhao, Wentian and Xin, Cheng and Wan, Kun and Yu, Lantao and Guo, Qianyu and Yu, Zixun and Lu, Yawen and others},
  booktitle={Proceedings of the IEEE/CVF Conference on Computer Vision and Pattern Recognition},
  pages={22160--22169},
  year={2024}
}

@inproceedings{yang2023freenerf,
  title={Freenerf: Improving few-shot neural rendering with free frequency regularization},
  author={Yang, Jiawei and Pavone, Marco and Wang, Yue},
  booktitle={Proceedings of the IEEE/CVF conference on computer vision and pattern recognition},
  pages={8254--8263},
  year={2023}
}

@inproceedings{somraj2023simplenerf,
  title={Simplenerf: Regularizing sparse input neural radiance fields with simpler solutions},
  author={Somraj, Nagabhushan and Karanayil, Adithyan and Soundararajan, Rajiv},
  booktitle={SIGGRAPH Asia 2023 Conference Papers},
  pages={1--11},
  year={2023}
}

@inproceedings{sargent2024zeronvs,
  title={Zeronvs: Zero-shot 360-degree view synthesis from a single image},
  author={Sargent, Kyle and Li, Zizhang and Shah, Tanmay and Herrmann, Charles and Yu, Hong-Xing and Zhang, Yunzhi and Chan, Eric Ryan and Lagun, Dmitry and Fei-Fei, Li and Sun, Deqing and others},
  booktitle={Proceedings of the IEEE/CVF Conference on Computer Vision and Pattern Recognition},
  pages={9420--9429},
  year={2024}
}

@inproceedings{deng2022depthnerf,
  title={Depth-supervised nerf: Fewer views and faster training for free},
  author={Deng, Kangle and Liu, Andrew and Zhu, Jun-Yan and Ramanan, Deva},
  booktitle={Proceedings of the IEEE/CVF conference on computer vision and pattern recognition},
  pages={12882--12891},
  year={2022}
}

@inproceedings{roessle2022densedepthnerf,
  title={Dense depth priors for neural radiance fields from sparse input views},
  author={Roessle, Barbara and Barron, Jonathan T and Mildenhall, Ben and Srinivasan, Pratul P and Nie{\ss}ner, Matthias},
  booktitle={Proceedings of the IEEE/CVF conference on computer vision and pattern recognition},
  pages={12892--12901},
  year={2022}
}

@inproceedings{wang2023sparsenerf,
  title={Sparsenerf: Distilling depth ranking for few-shot novel view synthesis},
  author={Wang, Guangcong and Chen, Zhaoxi and Loy, Chen Change and Liu, Ziwei},
  booktitle={Proceedings of the IEEE/CVF international conference on computer vision},
  pages={9065--9076},
  year={2023}
}

@inproceedings{niemeyer2022regnerf,
  title={Regnerf: Regularizing neural radiance fields for view synthesis from sparse inputs},
  author={Niemeyer, Michael and Barron, Jonathan T and Mildenhall, Ben and Sajjadi, Mehdi SM and Geiger, Andreas and Radwan, Noha},
  booktitle={Proceedings of the IEEE/CVF conference on computer vision and pattern recognition},
  pages={5480--5490},
  year={2022}
}

@inproceedings{jain2021dietnerf,
  title={Putting nerf on a diet: Semantically consistent few-shot view synthesis},
  author={Jain, Ajay and Tancik, Matthew and Abbeel, Pieter},
  booktitle={Proceedings of the IEEE/CVF international conference on computer vision},
  pages={5885--5894},
  year={2021}
}

@article{liu20243dgs,
  title={3dgs-enhancer: Enhancing unbounded 3d gaussian splatting with view-consistent 2d diffusion priors},
  author={Liu, Xi and Zhou, Chaoyi and Huang, Siyu},
  journal={Advances in Neural Information Processing Systems},
  volume={37},
  pages={133305--133327},
  year={2024}
}

@inproceedings{xu2025depthsplat,
  title={Depthsplat: Connecting gaussian splatting and depth},
  author={Xu, Haofei and Peng, Songyou and Wang, Fangjinhua and Blum, Hermann and Barath, Daniel and Geiger, Andreas and Pollefeys, Marc},
  booktitle={Proceedings of the Computer Vision and Pattern Recognition Conference},
  pages={16453--16463},
  year={2025}
}

@inproceedings{zhang2024fregs,
  title={Fregs: 3d gaussian splatting with progressive frequency regularization},
  author={Zhang, Jiahui and Zhan, Fangneng and Xu, Muyu and Lu, Shijian and Xing, Eric},
  booktitle={Proceedings of the IEEE/CVF Conference on Computer Vision and Pattern Recognition},
  pages={21424--21433},
  year={2024}
}

@inproceedings{li2024dngaussian,
  title={Dngaussian: Optimizing sparse-view 3d gaussian radiance fields with global-local depth normalization},
  author={Li, Jiahe and Zhang, Jiawei and Bai, Xiao and Zheng, Jin and Ning, Xin and Zhou, Jun and Gu, Lin},
  booktitle={Proceedings of the IEEE/CVF conference on computer vision and pattern recognition},
  pages={20775--20785},
  year={2024}
}

@inproceedings{zheng2025nexusgs,
  title={NexusGS: Sparse View Synthesis with Epipolar Depth Priors in 3D Gaussian Splatting},
  author={Zheng, Yulong and Jiang, Zicheng and He, Shengfeng and Sun, Yandu and Dong, Junyu and Zhang, Huaidong and Du, Yong},
  booktitle={Proceedings of the Computer Vision and Pattern Recognition Conference},
  pages={26800--26809},
  year={2025}
}

@inproceedings{xu2025dropoutgs,
  title={DropoutGS: Dropping Out Gaussians for Better Sparse-view Rendering},
  author={Xu, Yexing and Wang, Longguang and Chen, Minglin and Ao, Sheng and Li, Li and Guo, Yulan},
  booktitle={Proceedings of the Computer Vision and Pattern Recognition Conference},
  pages={701--710},
  year={2025}
}

@inproceedings{zhang2024corgs,
  title={Cor-gs: sparse-view 3d gaussian splatting via co-regularization},
  author={Zhang, Jiawei and Li, Jiahe and Yu, Xiaohan and Huang, Lei and Gu, Lin and Zheng, Jin and Bai, Xiao},
  booktitle={European Conference on Computer Vision},
  pages={335--352},
  year={2024},
  organization={Springer}
}

@inproceedings{zhao2025selfenblegs,
  title={Self-ensembling gaussian splatting for few-shot novel view synthesis},
  author={Zhao, Chen and Wang, Xuan and Zhang, Tong and Javed, Saqib and Salzmann, Mathieu},
  booktitle={Proceedings of the IEEE/CVF International Conference on Computer Vision},
  pages={4940--4950},
  year={2025}
}

@inproceedings{park2025dropgaussian,
  title={Dropgaussian: Structural regularization for sparse-view gaussian splatting},
  author={Park, Hyunwoo and Ryu, Gun and Kim, Wonjun},
  booktitle={Proceedings of the Computer Vision and Pattern Recognition Conference},
  pages={21600--21609},
  year={2025}
}

@article{lin2025depth,
  title={Depth anything 3: Recovering the visual space from any views},
  author={Lin, Haotong and Chen, Sili and Liew, Junhao and Chen, Donny Y and Li, Zhenyu and Shi, Guang and Feng, Jiashi and Kang, Bingyi},
  journal={arXiv preprint arXiv:2511.10647},
  year={2025}
}

@inproceedings{liu2023zero123,
  title={Zero-1-to-3: Zero-shot one image to 3d object},
  author={Liu, Ruoshi and Wu, Rundi and Van Hoorick, Basile and Tokmakov, Pavel and Zakharov, Sergey and Vondrick, Carl},
  booktitle={Proceedings of the IEEE/CVF international conference on computer vision},
  pages={9298--9309},
  year={2023}
}

@inproceedings{liu2024one2345,
  title={One-2-3-45++: Fast single image to 3d objects with consistent multi-view generation and 3d diffusion},
  author={Liu, Minghua and Shi, Ruoxi and Chen, Linghao and Zhang, Zhuoyang and Xu, Chao and Wei, Xinyue and Chen, Hansheng and Zeng, Chong and Gu, Jiayuan and Su, Hao},
  booktitle={Proceedings of the IEEE/CVF conference on computer vision and pattern recognition},
  pages={10072--10083},
  year={2024}
}

@article{shi2023zero123++,
  title={Zero123++: a single image to consistent multi-view diffusion base model},
  author={Shi, Ruoxi and Chen, Hansheng and Zhang, Zhuoyang and Liu, Minghua and Xu, Chao and Wei, Xinyue and Chen, Linghao and Zeng, Chong and Su, Hao},
  journal={arXiv preprint arXiv:2310.15110},
  year={2023}
}

@article{shi2023mvdream,
  title={Mvdream: Multi-view diffusion for 3d generation},
  author={Shi, Yichun and Wang, Peng and Ye, Jianglong and Long, Mai and Li, Kejie and Yang, Xiao},
  journal={arXiv preprint arXiv:2308.16512},
  year={2023}
}

@inproceedings{long2024wonder3d,
  title={Wonder3d: Single image to 3d using cross-domain diffusion},
  author={Long, Xiaoxiao and Guo, Yuan-Chen and Lin, Cheng and Liu, Yuan and Dou, Zhiyang and Liu, Lingjie and Ma, Yuexin and Zhang, Song-Hai and Habermann, Marc and Theobalt, Christian and others},
  booktitle={Proceedings of the IEEE/CVF conference on computer vision and pattern recognition},
  pages={9970--9980},
  year={2024}
}

@inproceedings{sauer2024adversarial,
  title={Adversarial diffusion distillation},
  author={Sauer, Axel and Lorenz, Dominik and Blattmann, Andreas and Rombach, Robin},
  booktitle={European Conference on Computer Vision},
  pages={87--103},
  year={2024},
  organization={Springer}
}

@inproceedings{reda2022film,
  title={Film: Frame interpolation for large motion},
  author={Reda, Fitsum and Kontkanen, Janne and Tabellion, Eric and Sun, Deqing and Pantofaru, Caroline and Curless, Brian},
  booktitle={European Conference on Computer Vision},
  pages={250--266},
  year={2022},
  organization={Springer}
}

@article{hirschorn2025splatent,
  title={Splatent: Splatting Diffusion Latents for Novel View Synthesis},
  author={Hirschorn, Or and Sela, Omer and Huberman-Spiegelglas, Inbar and Efrat, Netalee and Alshan, Eli and Ideses, Ianir and Devernay, Frederic and Zvik, Yochai and Fritz, Lior},
  journal={arXiv preprint arXiv:2512.09923},
  year={2025}
}

@inproceedings{niklaus2020softmaxsplatting,
  title={Softmax splatting for video frame interpolation},
  author={Niklaus, Simon and Liu, Feng},
  booktitle={Proceedings of the IEEE/CVF conference on computer vision and pattern recognition},
  pages={5437--5446},
  year={2020}
}

@inproceedings{chen2024mvsplat,
  title={Mvsplat: Efficient 3d gaussian splatting from sparse multi-view images},
  author={Chen, Yuedong and Xu, Haofei and Zheng, Chuanxia and Zhuang, Bohan and Pollefeys, Marc and Geiger, Andreas and Cham, Tat-Jen and Cai, Jianfei},
  booktitle={European Conference on Computer Vision},
  pages={370--386},
  year={2024},
  organization={Springer}
}

@inproceedings{charatan2024pixelsplat,
  title={pixelsplat: 3d gaussian splats from image pairs for scalable generalizable 3d reconstruction},
  author={Charatan, David and Li, Sizhe Lester and Tagliasacchi, Andrea and Sitzmann, Vincent},
  booktitle={Proceedings of the IEEE/CVF conference on computer vision and pattern recognition},
  pages={19457--19467},
  year={2024}
}

@inproceedings{patle2025ad,
  title={AD-GS: Alternating Densification for Sparse-Input 3D Gaussian Splatting},
  author={Patle, Gurutva and Girgaonkar, Nilay and Somraj, Nagabhushan and Soundararajan, Rajiv},
  booktitle={Proceedings of the SIGGRAPH Asia 2025 Conference Papers},
  pages={1--11},
  year={2025}
}
